\documentclass[10pt, a4paper]{article}

\usepackage{lrec2026} 
\usepackage{tikz}
\usetikzlibrary{arrows.meta,positioning,fit,calc,shapes.multipart}

\title{Integrating Knowledge Graphs and Multilingual Scholarly Corpora for Domain-Adaptive LLMs in SSH}

\name{\parbox{\textwidth}{\centering
Adam Faci\textsuperscript{1}, Alessio Miaschi\textsuperscript{2}, Anne Combe\textsuperscript{3}, Pascal Cuxac\textsuperscript{4}, \\Francesca Frontini\textsuperscript{2},
Nicolas Larrousse\textsuperscript{1}, Stéphane Pouyllau\textsuperscript{1}
}}

\address{\textsuperscript{1} Huma-Num, CNRS, Paris, France \\
\textsuperscript{2}CNR - Institute for Computational Linguistics ``A. Zampolli'' (CNR-ILC), Pisa, Italy \\
\textsuperscript{3}Inria, France \\
\textsuperscript{4}Inist, CNRS, 2 rue Jean Zay, 54500 Vandoeuvre-lès-Nancy, France \\
    \{name.surname\}@huma-num.fr, 
    \{name.surname\}@ilc.cnr.it \\
    anne.combe@inria.fr, pascal.cuxac@inist.fr
      }

\abstract{
The integration of Large Language Models (LLMs) into scientific research workflows, particularly for bibliographic discovery and literature synthesis, raises significant methodological, epistemic and regulatory challenges for the Social Sciences and Humanities (SSH), especially with regard to disciplinary diversity, multilingual access to sources and the evaluation of results. This paper presents an on-going use case developed within the European project LLMs4EU and the ALT-EDIC infrastructure, aimed at adapting foundation models to SSH research practices and supporting tasks such as question answering, comparative document analysis and literature review. The evaluation framework follows the LLMs4EU protocol and encompasses both independent quantitative benchmarking (retrieval, summarisation, traceability and hallucination detection) and a qualitative assessment involving a panel of Digital Humanities experts. By embedding model adaptation within research infrastructures and a structured legal and ethical compliance framework, the use case explores how domain-sensitive and regulation-aware generative AI can support SSH scholarship while preserving reliability and epistemic responsibility.
\\ \newline  }

\begin{document}

\maketitleabstract

\section{Introduction}

In recent years, Large Language Models (LLMs) have profoundly transformed the landscape of language technologies and research practices in Social Sciences and Humanities  (SSH - see, among possible examples, \citealt{arachchige_proceedings_2025}). At the same time, the development of both LLMs and AI-enhanced scientific discovery platforms remains strongly concentrated on dominant languages and cultural contexts. It also tends to privilege epistemological models and validation practices that are typical of the so-called “hard” sciences.

This imbalance is reflected in many established and emerging AI-powered platforms for exploring scientific literature. Semantic search engines (e.g. Semantic Scholar), AI-assisted discovery tools (e.g. Elicit), large-scale citation indexes (e.g. Scopus and Web of Science), and conversational systems integrated into publishing ecosystems (e.g. Scite Assistant) are predominantly trained on large corpora of English-language journal articles. In most cases, they allow querying primarily or exclusively in English, prioritise peer-reviewed journal publications over other forms of scholarly output (such as monographs, critical editions, or research data), and rely heavily on citation-based metrics—such as impact factor or h-index—to rank and filter results. 

Such design choices are not neutral. They reflect and reinforce specific models of scientific production, visibility, and authority that may not align with the diversity of practices in SSH, where multilingualism, book-based scholarship, local research traditions, and qualitative evaluation criteria remain central. As a consequence, SSH scholars working in less-resourced languages or within epistemic traditions that do not conform to citation-driven models risk being further marginalised in the emerging AI-mediated research ecosystem.

Moreover, research in the SSH often relies on interpretation and contextualisation. Texts, data, and metadata are not treated as fixed or neutral objects. Their meaning depends on research questions, theoretical perspectives, and historical context. The same material can therefore be read and organised in different ways. A well-known author such as Dante offers a simple example. A literary scholar may focus on narrative structure, poetic language, and intertextual references in the \textit{Commedia}. A philologist may study manuscript traditions, textual variants, and editorial history. A historian may examine Dante’s political thought in relation to the conflicts of medieval Italy.  A Digital Humanities scholar, in turn, might build a digital scholarly edition, encode the text in TEI-XML, compare manuscript witnesses computationally, or analyse the circulation of Dante’s works through network analysis and corpus-based methods. In such contexts, an algorithm cannot rely on topic matching alone, but should also recognise methodological affinities, disciplinary perspectives, and also research methodologies.

To address these challenges, Europe has developed a rich ecosystem of  data repositories and publication platforms designed to preserve the epistemic and methodological specificities of SSH. Cross-border disciplinary infrastructures such as CLARIN ERIC, DARIAH ERIC, and OPERAS have been conceived to support the shared use of digital resources while maintaining strong attention to disciplinary traditions and linguistic diversity \cite{branco_clarin_2023,dumouchel_triple_2020,konig_ssh_2023}. These infrastructures have developed dedicated discovery services and meta-catalogues—such as the CLARIN Virtual Language Observatory, the SSH Open Marketplace, and the GoTriple platform—explicitly tailored to the needs of SSH communities. Such environments provide an important alternative to commercial discovery tools and could benefit from enhanced functionalities powered by LLMs. At the national level, infrastructures such as the French Huma-Num offer search facilities including Nakala and ISIDORE, which are widely used by humanities researchers\footnote{\url{https://www.huma-num.fr/les-services-par-etapes/}}.

Research has shown that these digital environments are not neutral repositories, but are shaped around the practices and expectations of their scholarly communities. More recent studies on research infrastructures and data archiving services further emphasise that humanities data emerge from dynamic collective processes, in which archives and service providers act not only as technical support structures, but also as mediators and co-constructors of research communities  \cite{morselli_fostering_2025}. This perspective reinforces the idea that tools for discovery and analysis must remain sensitive to disciplinary practices rather than imposing external models of relevance.

Another distinctive feature of humanities research is the plurality of scholarly outputs and the central role of multilingualism as an epistemic and methodological dimension, as extensively discussed in multilingual Digital Humanities studies \cite{balula2021multilingualism,viola_multilingual_2024}. Alongside journal articles, knowledge production includes datasets, curated corpora, digital editions, workflows, and forms of scholarly communication such as research blogs. The relevance of these outputs—and the need for appropriate citation, indexing, and valorisation practices—has been widely documented \cite{barbot_contextualizing_2024,mayeur_imparting_2017}. Any AI-enhanced discovery system that aims to serve SSH communities must therefore account for this diversity of formats, languages, and publication cultures.

In the context outlined above, the integration of LLM-based functionalities for bibliographic and documentary research in SSH cannot be reduced to generic applications. Language models need to be designed and evaluated in ways that reflect the specific research practices of the humanities, support interdisciplinary and transdisciplinary perspectives, and facilitate meaningful interaction among scholarly communities. The use case we propose,  \textit{ReSearch\_SSH},  is a collaboration between French and Italian SSH research infrastructures within the LLMs4EU (Large Language Models for the European Union) project, funded by the European Commission and coordinated by ALT-EDIC (Alliance for Language Technologies – European Digital Infrastructure Consortium)\footnote{\url{https://www.alt-edic.eu/}}.  It aims to enhance an existing search platform by integrating LLM-powered functionalities for advanced retrieval and for supporting the construction of state-of-the-art overviews grounded in curated SSH catalogs.

As noted by \citet{fenlon_thematic_2017}, Digital Humanities (DH) has witnessed the emergence of new forms of scholarly production, including research blogs, large-scale digital corpora, tools, and platforms that challenge traditional models of academic publishing. These outputs are well represented in the infrastructures mentioned above, and DH itself is a particularly dynamic research area, characterised by strong multilingualism and interdisciplinarity. For these reasons, DH constitutes a compelling testbed for the integration of LLM-based search functionalities. DH scholars, accustomed to working across methods, languages, and formats, are well positioned to act as expert users in the evaluation and co-construction of these new tools.

This paper is organized as follows. After framing the role of LLMs from the perspective of SSH research practices, we introduce the European LLMs4EU project within the broader context of the new ALT-EDIC infrastructure. We then present the ReSearch\_SSH use case as a case study of model adaptation and fine-tuning for the SSH domain, focusing on data resources, methodology and finetuning strategies, and lastly on the foreseen evaluation strategies and legal issues. Finally, we suggest the involvement of research communities, especially within Digital Humanities, through the establishment of an expert panel, outlining possible forms of participation of the SSH communities in the development and evaluation of LLMs for research support.


\section{ALT-EDIC and LLMs4EU}

ALT-EDIC is a European Digital Infrastructure Consortium, a legal framework designed to enable several Member States to jointly develop and operate strategic digital infrastructures of common European interest.  Specifically, ALT-EDIC aims to support European excellence in language technologies and to promote European linguistic diversity, following a cooperative model that brings together public institutions, industry, civil society and research. This is reflected in its close collaboration with CLARIN ERIC, the research infrastructure for language resources, and with several of its national nodes.

Established in 2024, ALT-EDIC carries out its activities through participation in several complementary European projects. Among these, OpenEuroLLM focuses on the development of open and multilingual language models, representing a key component of the European strategy for open, transparent, and value-aligned Artificial Intelligence.

Alongside this first pillar, the LLMs4EU project addresses the more applied dimension of Large Language Models across strategic sectors, adopting a strongly use-case-driven approach. The project is structured around five application domains, i.e. tourism, public services, telecommunications, energy, and science, and focuses on the fine-tuning and adaptation of open models to concrete contexts and needs. These use cases are not conceived as simple application demonstrators, but rather as experimental laboratories in which model adaptation requirements are defined, domain data are selected and prepared, fine-tuning and task-tuning strategies are tested, and evaluation protocols are designed. The overarching goal is to anchor the development of LLMs in real-world needs and in concrete communities of practice.

\section{The \textit{ReSearch\_SSH} use case}

Of particular interest to the European Digital Humanities community is the \textit{ReSearch\_SSH} use case, which belongs to the \textit{Science} domain of the LLMs4EU project. It focuses on the development and experimental evaluation of generative language models adapted to advanced tasks for scholarly discovery and synthesis of scientific literature in the Social Sciences and Humanities. Particular attention is devoted to factuality, source traceability, and explainability of results. The initiative is grounded in close collaboration between major French and Italian research infrastructures, 

Rather than building a new discovery system, the use case extends the existing ISIDORE platform\footnote{\url{https://isidore-project.eu/}}, a widely used French search engine for humanities and social sciences (SSH) that aggregates multilingual scholarly resources. ISIDORE thus functions both as a large-scale documentary infrastructure and as an operational search environment.

The system follows a retrieval-augmented generation (RAG) architecture. Its retrieval layer is based on a domain-adapted model trained on historical ISIDORE queries, aligning retrieval with actual SSH search behaviour. This component is strengthened through graph-based techniques: contextualisation relies mainly on the Wikidata knowledge graph for entity disambiguation and semantic expansion, while relational structures derived from metadata support structured navigation across authors, themes, institutions, and citations. Document-level semantic graphs further enable fine-grained comparison between research outputs.

A multilingual large language model forms the generative layer, operating over the corpus indexed within ISIDORE and informed by metadata and graph-based signals. It supports natural-language querying, assisted literature synthesis, and the production of structured overviews grounded in explicitly retrieved sources. By combining adaptive retrieval with graph-aware contextualisation and generative reasoning, the system supports core SSH workflows such as literature review, thematic exploration, and comparative analysis, while ensuring that all generated outputs remain transparent and traceable to indexed documents.

The project is currently ongoing. In its present phase, it concentrates on enabling advanced querying in Italian and English over predominantly French scholarly materials aggregated within the ISIDORE ecosystem, as an initial step toward broader multilingual coverage.

Operationally, the use case is structured along four main stages. 

\begin{itemize}
    \item First, a phase of domain alignment and multilingual enrichment adapts the base model to SSH discourse using large-scale curated corpora.
    \item Second, retrieval and representation components are optimised in a retrieval-augmented generation (RAG) framework, leveraging ISIDORE’s metadata and indexing infrastructure. 
    \item Third, the model is fine-tuned for downstream scholarly tasks, including multilingual document retrieval, literature synthesis, and structured state-of-the-art construction. 
    \item Finally, the system is evaluated through expert-based validation protocols designed to assess relevance, factual grounding, and usability in realistic SSH research scenarios.

\end{itemize}

In what follows we will illustrate the datasets, finetuning plan, and evaluation strategy.

\subsection{Data and disciplinary focus}

The primary corpus for large-scale domain alignment is the SSH subset of the ISTEX database\footnote{\url{https://www.istex.fr/ressources-pour-le-fouille-de-textes/}}. ISTEX aggregates more than 30 million scientific documents and provides full texts in XML/TEI format together with standardised and enriched metadata \cite{de_salabert_vers_2020}. For the purposes of ReSearch\_SSH, an SSH-specific extraction (approximately 3 million documents) has been prepared based on disciplinary classifications and quality criteria (see Table \ref{tab:istex}).

Full texts are available in TEI/XML; metadata are provided in JSON format and include title, abstract, authors, affiliations, publisher, year, structured references, scientific classifications, and named entities.

\begin{table}[h]
\centering
\begin{tabular}{lr}
\hline
Total documents & $\sim$ 3,000,000 \\
Languages represented & 60+ \\
Full-text format & XML/TEI \\
Metadata license & Etalab (open) \\
\hline
\end{tabular}
\caption{ISTEX SSH subset overview.}
\label{tab:istex}
\end{table}

\begin{table}[h]
\centering
\begin{tabular}{lr}
\hline
Language & Tokens (millions, approx.) \\
\hline
English & 16,000–21,000 \\
French & 450–600 \\
Portuguese & 110–150 \\
German & 40–55 \\
Spanish & 40–50 \\
Italian & 10–15 \\
\hline
\end{tabular}
\caption{Estimated token distribution in ISTEX SSH subset.}
\label{tab:istex_stats}
\end{table}

Although English dominates in volume, French constitutes the main non-English component, which is particularly relevant for the ISIDORE ecosystem. The corpus therefore supports large-scale alignment to SSH discourse while preserving multilingual exposure (see Table \ref{tab:istex_stats}).

To complement the French-dominant SSH material, targeted Italian and bilingual DH datasets are incorporated to strengthen multilingual and disciplinary alignment. These encompass the Italian Digital Humanities research papers from the annual AIUCD conference series and the Peer-reviewed Italian journal in Digital Humanities (266 articles, CC-BY license) (Tables \ref{tab:aiucd} and \ref{tab:ud}).

\begin{table}[h]
\centering
\begin{tabular}{lr}
\hline
Language & Tokens (millions, approx.) \\
\hline
English & 1.2 \\
Italian & 0.8 \\
\hline
Total & 2 \\
\hline
\end{tabular}
\caption{AIUCD Proceedings token distribution.}
\label{tab:aiucd}
\end{table}

\begin{table}[h]
\centering
\begin{tabular}{lr}
\hline
Language & Tokens (millions, approx.) \\
\hline
English & 1.1 \\
Italian & 0.9 \\
\hline
Total & 1 \\
\hline
\end{tabular}
\caption{Umanistica Digitale token distribution.}
\label{tab:ud}
\end{table}

These corpora provide high-quality DH-specific discourse, enabling: i) genre adaptation (conference papers, journal articles); ii) alignment with Italian DH terminology; and iii) cross-lingual bridging between Italian, French and English SSH traditions.

On top of this generalist alignment layer, the use case may introduce also scientific publications and academic blog posts from the Hypotheses\footnote{\url{https://hypotheses.org/}} platform, together with research data and metadata from the Nakala repository\footnote{\url{https://www.nakala.fr/}}, will be incorporated to expose the models to DH-specific practices, formats and epistemic communities.

\subsection{Methodological approach and fine-tuning}

The proposed approach adopts a Knowledge-Graph-based Retrieval-Augmented Generation architecture (GraphRAG; see, among others, \citealt{edge2024local}), designed to ensure that generated answers are explicitly grounded in the most relevant reference documents and that this grounding remains interpretable.

Model adaptation is organised in complementary stages. A first phase performs domain alignment through continued pre-training on SSH corpora in French and Italian, strengthening linguistic and terminological competence while preserving the general reasoning capacities of the base multilingual model.

A second phase distinguishes retrieval optimisation from generative task adaptation. The retrieval component is fine-tuned on historical query–interaction data from ISIDORE. Training signals are derived from user behaviour: documents effectively clicked and consulted are promoted, while systematically ignored results are ranked lower. This interaction-aware strategy aligns ranking with observed scholarly practices rather than relying solely on textual similarity.

In parallel, the generative component is instruction-tuned for research-oriented tasks within the GraphRAG pipeline. Particular emphasis is placed on the construction and extension of state-of-the-art overviews. Fine-tuning relies on examples of surveys and literature reviews combined with graph-based contextual signals encoding thematic, citation, and authorship relations. The objective is to enable the model to organise retrieved materials into coherent research syntheses rather than merely producing isolated summaries.

A central design constraint concerns query understanding. The system relies on a medium-scale multilingual model rather than very large reasoning-focused architectures, making robust interpretation of exploratory research queries essential. Fine-tuning therefore integrates panel-based analyses of user search practices in SSH, capturing how scholars formulate evolving and open-ended information needs. 

The generation of synthetic fine-tuning data for the query understanding component will be investigated through multiple strategies, comparing both commercial models (e.g., GPT-5) and open European language models. Given the project's alignment with the OpenEuroLLM initiative and with the broader ALT-EDIC objective of promoting sovereign and transparent AI infrastructures, strong preference is given to open European models. Among these, the EuroLLM family \cite{martins2025eurollm}, given their strong multilingual capabilities and open licensing terms. Closed commercial models will be used primarily as reference baselines to assess whether comparable data quality can be achieved through fully open pipelines, with the explicit goal of converging on a production workflow that operates entirely on open and auditable models, ensuring reproducibility and compliance with the project's governance principles.
More broadly, the base models selected for domain-adaptive fine-tuning within the ReSearch\_SSH use case will be drawn from the pool of open multilingual foundation models developed or endorsed within the European AI ecosystem. The final selection will be determined on the basis of multilingual coverage (with particular attention to French, Italian, and English), licensing compatibility with the intended research deployment, and baseline performance on SSH-relevant tasks. Specific model identifiers and version details will be reported in conjunction with the first experimental results.

Knowledge graphs — including those provided by Wikidata and OpenAIRE \cite{manghi_openaire_2019}, as well as the metadata and relational structures of ISIDORE and Nakala — enrich retrieval, strengthen document linking, and enhance the transparency of generated results. Structured intermediate representations may support internal reasoning, while explicit source attribution ensures traceability for end users.

An important consideration concerns the portability of the proposed architecture beyond the specific infrastructure on which it is developed. The GraphRAG pipeline can be decomposed into components with different degrees of infrastructure dependency. The core retrieval-augmented generation mechanism - comprising a retrieval model, a reranking layer, and a generative module operating over retrieved passages — is infrastructure-agnostic and can function with any document collection exposed through standard indexing interfaces. Similarly, the knowledge graph enrichment layer relies primarily on publicly available resources such as Wikidata and OpenAIRE, which are not tied to specific institutional environments and can be integrated into diverse system architectures. The most infrastructure-specific component is the interaction-aware retrieval fine-tuning, which exploits historical query–click data from ISIDORE. In the absence of comparable behavioural signals, this component can be replaced by purely semantic retrieval models or by lightweight relevance feedback mechanisms. Likewise, while the availability of full texts in TEI/XML format facilitates fine-grained document parsing and section-level retrieval, the system does not strictly require this format: plain-text corpora accompanied by structured metadata in standard formats (e.g., JSON, Dublin Core) can serve as an adequate alternative, albeit with some reduction in retrieval granularity.

\subsection{Deployment and evaluation}

The fine-tuned models will be deployed within a sandbox environment of the ISIDORE platform, conceived as an experimental ``ISIDORE AI'' workspace in which advanced research assistance functionalities can be tested in a controlled setting. The system will be integrated with the ISIDORE databases and knowledge graphs and will allow users to formulate complex queries and obtain organised and synthesised results accompanied by explicit references to the underlying sources. Multilingual querying will also be supported, potentially through the integration of a dedicated translation model.

System evaluation will be conducted in accordance with the LLMs4EU evaluation protocol, which mandates the involvement of an independent team external to the use case implementation. This team is responsible for defining realistic and scenario-based evaluation procedures grounded in the specific use case requirements and fine-tuning strategies. The independent evaluation plan, currently under development, will include quantitative assessment of retrieval performance, multi-document summarization quality, source traceability, and hallucination detection. Particular attention will be devoted to cross-lingual performance, reflecting the multilingual nature of the use case. The evaluation process may require the construction of ad hoc benchmark datasets tailored to SSH research scenarios.

A central component of the evaluation framework will consist of a structured qualitative assessment conducted by expert panels drawn from the French and Italian Digital Humanities communities. These panels will evaluate system outputs in terms of scholarly quality, epistemic reliability, methodological adequacy, and practical usefulness for research workflows. Beyond assessment, panel feedback will also inform preference optimisation processes (e.g., alignment tuning approaches), contributing to the refinement of the models so that they better reflect disciplinary standards, interpretative practices, and relevance criteria specific to Digital Humanities research.

\section{Legal and regulatory compliance}

ReSearch\_SSH is designed in full compliance with the European regulatory framework on Artificial Intelligence, data protection, and intellectual property. In line with the governance structure of LLMs4EU, the use case is supported by both a Data Management Plan and a dedicated Legal and Ethics Compliance Assessment Plan, developed in coordination with the Legal and Ethics work package of the project. This structured governance framework ensures that data selection, model adaptation strategies, and deployment modalities are continuously assessed in light of applicable European legislation and of the principles of lawfulness, responsibility, transparency and proportionality guiding the project.

With respect to the EU AI Act, the system is conceived as a downstream modification of an existing general-purpose AI (GPAI) model. The use case does not qualify as a high-risk AI system as it operates as a research assistance tool without automated decision-making affecting individual rights. Nevertheless, transparency, documentation, and traceability requirements are treated as central design principles. In particular, the adoption of a KG-enhanced RAG architecture enforces source grounding and explicit citation mechanisms, thereby mitigating hallucination risks and strengthening epistemic reliability. Deployment is limited to a research-oriented sandbox environment within the ISIDORE infrastructure, and no unrestricted public release of licensed full texts or training data is foreseen. Human oversight is structurally embedded: system outputs are advisory, challengeable, and subject to expert validation.

From the perspective of personal data protection, the project operates in accordance with the General Data Protection Regulation (GDPR). Personal data may appear in scholarly corpora (e.g., author names, affiliations, citation metadata) and are processed only insofar as necessary for the purpose of the use case and to ensure scientific attribution, source traceability and hallucination mitigation. These data are retained until the end of the use case and are not used for profiling or behavioural modelling. Evaluation activities involving expert panels rely on informed consent procedures; panel data are limited to professional identifiers (e.g., institutional affiliation, disciplinary area) and are strictly separated from training data. In accordance with the principles of data minimisation and purpose limitation, panel data are used exclusively for evaluation and system validation, and are anonymised or deleted after the completion of the evaluation phase, though they may be made available in trusted repositories in aggregated and anonymised form.

Particular attention is devoted to copyright and database rights. The use case relies on heterogeneous data sources, including: (i) open-access SSH publications (e.g., under CC-BY or CC-BY-SA licences), (ii) metadata infrastructures released under open licences (e.g., Etalab), and (iii) licensed full texts from ISTEX, accessed under specific contractual conditions and Non-Disclosure Agreements. Licensed full texts are processed within controlled institutional environments, and retrieval-based responses are grounded in citation and excerpt linking mechanisms. The project acknowledges that the availability of content under open licences does not automatically imply unrestricted reusability for model fine-tuning, particularly when uses extend beyond the original dissemination purposes. In line with recent discussions (e.g., \cite{spichtinger_perspective_2026}), tensions between open science principles and AI training practices require case-by-case legal assessment, including consideration of text and data mining (TDM) exceptions under Directive (EU) 2019/790.

Overall, ReSearch\_SSH adopts a compliance-by-design approach, integrating legal analysis, risk assessment, and mitigation measures throughout the development lifecycle. This includes secure institutional storage, controlled access to licensed datasets, separation of personal identifiers from evaluation content, documentation of compute resources, bias and hallucination monitoring in coordination with the Evaluation team of the project, and explicit user disclaimers at deployment. This model reflects the broader LLMs4EU objective of developing multilingual language technologies that are not only performant, but also legally robust, ethically grounded, and aligned with European values.

\section{Current state of the use case}

The ReSearch\_SSH use case is currently in an advanced preparatory phase. The multilingual corpus is being consolidated through the selection, harmonisation and verification of SSH datasets, including licensed and open-access resources. Subsequently,  attention will devoted to metadata normalisation, knowledge graph integration and the preparation of domain-aligned subsets suitable for model adaptation. In parallel, the fine-tuning strategy, the evaluation framework and the Legal and Ethics Compliance Plan are being finalised in coordination with the relevant LLMs4EU work packages.

The first stages of domain alignment and retrieval-oriented fine-tuning are scheduled to begin shortly. These initial experiments will focus on controlled adaptation of an open foundation model to SSH scholarly discourse and on the optimisation of the KG-RAG pipeline. At this stage, particular care is taken to ensure that technical development proceeds in parallel with the independent evaluation protocol defined within LLMs4EU. 

A central element currently under development is of course the constitution of an expert panel composed of scholars and professionals from the Social Sciences and Humanities, with a specific focus on Digital Humanities communities in France and Italy. Crucially evaluation activities will  address not only model outputs, but the overall epistemic adequacy of the system in DH contexts. 

The panel may also support experimentation with human-in-the-loop approaches and with evaluation tasks that go beyond generic benchmarks, privileging complex and context-sensitive research scenarios typical of humanities scholarship.  From this perspective, the involvement of Digital Humanities communities aims to guide decisions related to data selection, annotation strategies and evaluation tasks, fostering the integration of disciplinary corpora that are representative of textual genres and research practices in the field.

\section{Conclusions}

This paper has presented ReSearch\_SSH, a use case developed within the LLMs4EU project in the Science domain, aimed at adapting large language models to the specific epistemic and methodological requirements of Social Sciences and Humanities research. Rather than proposing yet another generic conversational interface, the initiative explores how domain-aligned models, integrated within a knowledge-graph-enhanced RAG architecture and deployed inside an existing research infrastructure (ISIDORE), can support scholarly discovery while preserving traceability, multilingualism and disciplinary specificity.

The approach adopted in ReSearch\_SSH is grounded in three interconnected principles. First, adaptation is domain-driven: model alignment is guided by SSH corpora, research practices and multilingual requirements rather than by generic performance benchmarks. Second, evaluation is hybrid and independent: quantitative metrics are complemented by expert panel assessment, and evaluation protocols are defined by a team external to the use case implementation to ensure methodological robustness. Third, compliance is embedded by design: legal, regulatory and ethical considerations—ranging from GDPR and copyright to AI Act transparency obligations—are treated not as ex post constraints but as structural components of the system architecture and governance model.

At its current stage, the use case is moving from corpus consolidation and governance planning toward initial fine-tuning experiments and the constitution of a Franco-Italian expert panel. This transition marks a shift from conceptual design to empirical validation. The forthcoming phases will test whether the integration of domain adaptation, Graph-RAG grounding and participatory evaluation can effectively produce models that are not only technically performant, but also epistemically trustworthy and legally robust.

More broadly, ReSearch\_SSH contributes to an emerging reflection within Digital Humanities on the role of large language models as infrastructural components of research ecosystems. By situating LLM adaptation within European open science infrastructures and within a coordinated legal and evaluation framework, the project aims to demonstrate that multilingual, domain-sensitive and regulation-compliant generative systems are both technically feasible and methodologically desirable. While the current implementation is anchored to a specific infrastructure, the modular design of the architecture is intended to facilitate transferability. Future work will investigate the deployment of individual components in institutional settings with varying levels of metadata quality and corpus structure, with the aim of identifying minimal infrastructure requirements for effective SSH-oriented language model support.

\section{Acknowledgments}
\label{sec:ack}
This work has been supported by \textit{LLMs4EU ``Large Language Models for the European Union''} project, funded by the European Union through the Digital Europe Programme (DIGITAL-2024-AI-B-06-LANGUAGE - GA 101198470) under the grant agreement 101198470.

\section{Bibliographical References}\label{sec:reference}

\bibliographystyle{lrec2026-natbib}
\bibliography{lrec2026-example}

\end{document}